\documentclass[twoside,leqno,twocolumn]{article}

\usepackage[letterpaper]{geometry}

\usepackage{ltexpprt}
\usepackage{hyperref}

\usepackage{amsfonts}
\usepackage{amsmath}
\usepackage{amssymb}
\usepackage{color}
\usepackage{booktabs}
\usepackage{multirow}
\usepackage{ulem}
\usepackage{graphicx}
\usepackage{subfigure}
\usepackage{cleveref}
\usepackage{paralist}

\def\our{VisTabNet}
\def\R{\mathbb{R}}

\begin{document}

\title{\Large VisTabNet: Adapting Vision Transformers for Tabular Data
}
\author{Witold Wydmański\thanks{Faculty of Mathematics and Computer Science, Jagiellonian University, Kraków, Poland}\hskip 5pt\thanks{wwydmanski@gmail.com} \and Ulvi Movsum-zada$^*$ \and Jacek Tabor$^*$ \and Marek \'Smieja$^*$\thanks{marek.smieja@uj.edu.pl}}

\date{}

\maketitle


\fancyfoot[R]{\scriptsize{Submitted to SIAM Conference on Data Mining}}





\begin{abstract} \small\baselineskip=9pt 
Although deep learning models have had great success in natural language processing and computer vision, we do not observe comparable improvements in the case of tabular data, which is still the most common data type used in biological, industrial and financial applications. In particular, it is challenging to transfer large-scale pre-trained models to downstream tasks defined on small tabular datasets. To address this, we propose \our{} -- a cross-modal transfer learning method, which allows for adapting Vision Transformer (ViT) with pre-trained weights to process tabular data. By projecting tabular inputs to patch embeddings acceptable by ViT, we can directly apply a pre-trained Transformer Encoder to tabular inputs. This approach eliminates the conceptual cost of designing a suitable architecture for processing tabular data, while reducing the computational cost of training the model from scratch. Experimental results on multiple small tabular datasets (less than 1k samples) demonstrate VisTabNet's superiority, outperforming both traditional ensemble methods and recent deep learning models.
The proposed method goes beyond conventional transfer learning practice and shows that pre-trained image models can be transferred to solve tabular problems, extending the boundaries of transfer learning. We share our example implementation as a GitHub repository available at \url{https://github.com/wwydmanski/VisTabNet}.
\end{abstract}

\section{Introduction}

Deep learning has achieved tremendous success in various domains, including natural language processing (NLP) \cite{young_recent_2018}, computer vision (CV) \cite{krizhevsky_imagenet_2017}, and reinforcement learning (RL) \cite{atari}. Transformers, in particular, have become one of the most prominent neural architectures in NLP \cite{vaswani_attention_2023} and CV \cite{dosovitskiy_image_2021}, demonstrating remarkable improvements through their self-attention mechanism that captures global dependencies across inputs. In consequence, it is not surprising that several works focus on applying transformers beyond CV and NLP domains.

In real-world applications, tabular data remains one of the most common data types, used extensively in biology \cite{soueidan2015machine}, medicine \cite{rajkomar2019machine}, finance \cite{chowdhury2020predicting}, and manufacturing \cite{park2019reinforcement}. Recent reports indicate that data science practitioners work with tabular data as frequently as with texts or images\footnote{\url{https://www.statista.com/statistics/1241924/worldwide-software-developer-data-uses/}}. This is reflected in Kaggle's dataset distribution\footnote{Statistics gatherer in 2023 from \url{https://www.kaggle.com/datasets}}, where 6,688 datasets are tagged as "tabular", compared to 4,908 tagged as "image" and 178 as "text".

Despite the prevalence of tabular data in practical applications, deep learning models have yet to demonstrate significant improvements over traditional ensemble methods like XGBoost and Random Forests in this domain \cite{Chen2016XGBoostAS, kossen2021self}. The heterogeneous nature of tabular data and typically small sample sizes pose particular challenges for deep learning approaches \cite{grinsztajn2022tree, mcelfresh2024neural}. While various deep learning approaches, including transformer architectures \cite{gorishniy_revisiting_2021} and hypernetworks \cite{wydmanski_hypertab_2023}, have been adapted for tabular data, pre-training and transferring these models to downstream tasks remains challenging \cite{zhu2023xtab}.

\begin{figure}[!htb]
\includegraphics[width=\linewidth]{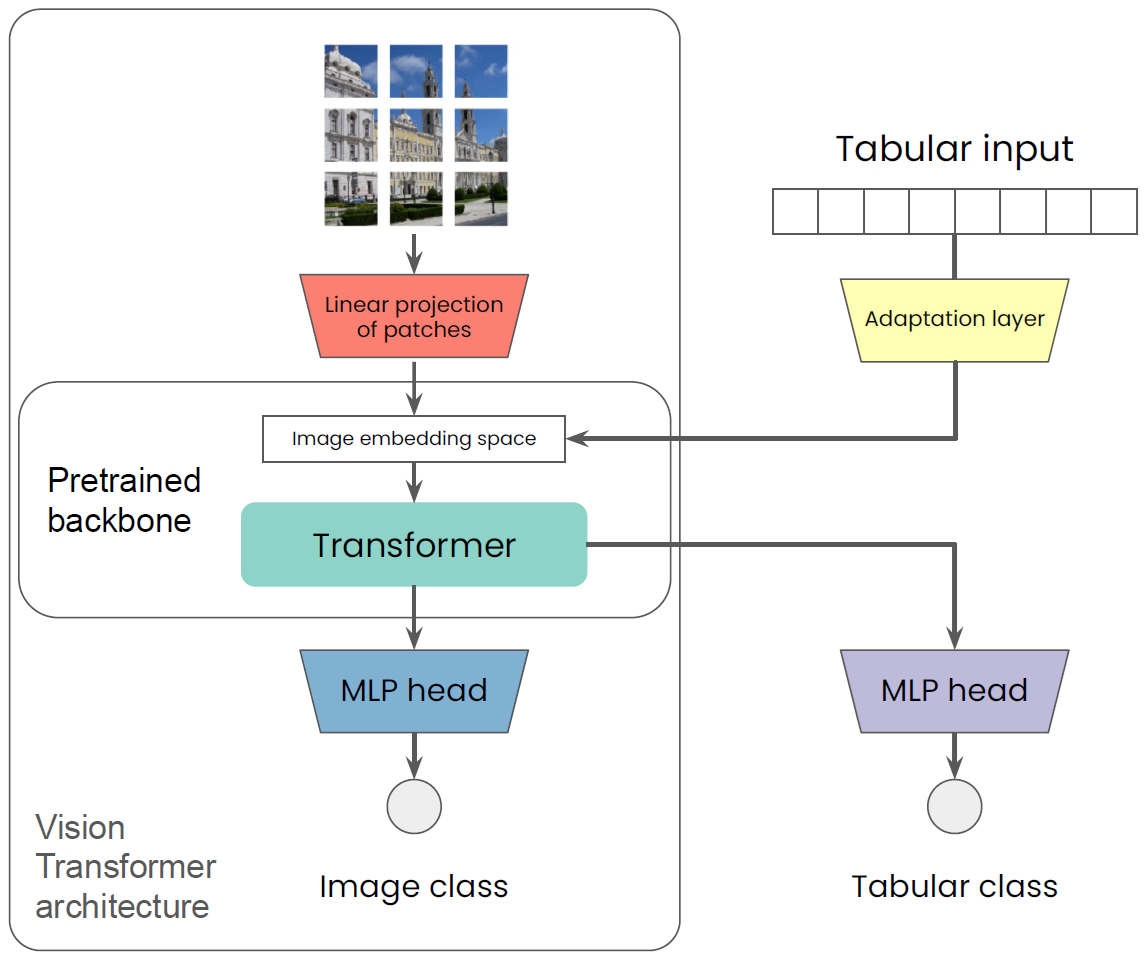}
\caption{Data flow architecture in VisTabNet. The tabular input is transformed into the image embedding space via our adaptation layer. After processing with pre-trained Transformer, the data is then classified using an MLP head. \label{fig:vistabnet}}
\centering
\end{figure}

In this paper, we propose a novel approach to cross-modal transfer learning, reusing the Vision Transformer (ViT) for tabular data tasks; see \Cref{fig:vistabnet}. In contrast to the typical reasoning behind transferring a feature extractor inside the same domain, we explore cross-modal transfer. More precisely, we take the Transformer Encoder of ViT pre-trained on image data and introduce an adaptation network that maps tabular inputs to a form compatible with the pre-trained ViT Encoder, enabling the construction of meaningful representations while avoiding the computational cost of training from scratch. Our \our{} model demonstrates superior performance across multiple conventional tabular datasets (\Cref{tab:benchmark}) and shows effective few-shot transfer capabilities (\Cref{fig:few}). Through extensive experimental analysis (\Cref{sec:analysis}), we provide insights into the effectiveness of cross-modal transfer and practical applications of \our{}.

Our contributions are summarized as follows:
\begin{compactitem}
    \item We propose a novel idea for cross-modal transfer learning, enabling the use of intrinsic patterns from one data modality to enhance training efficiency in another.
    \item We introduce VisTabNet, a novel deep learning algorithm that leverages the middle layers of a Vision Transformer architecture to process tabular data.
    \item We perform a comprehensive benchmark of VisTabNet against widely used shallow and deep learning methods across multiple diverse datasets, demonstrating its performance in learning from small datasets.
\end{compactitem}

\section{Related Work}

Tabular data is one of the most prevalent mediums in the world, right next to natural language, with over 5400 datasets present in OpenML~\cite{Vanschoren2014OpenMLNS} alone. For comparison, the most common NLP task in the Huggingface Dataset~\cite{lhoest-etal-2021-datasets} repository, text classification, is present as a tag in just 2300 datasets.

In contrast to computer vision or natural language processing, shallow models, such as Support Vector Machines ~\cite{cortes1995support}, Random Forests ~\cite{Breiman2001RandomF} and Gradient Boosting ~\cite{Friedman2001GreedyFA}, are usually the first choice for learning from tabular datasets. In particular, the family of Gradient Boosting algorithms ~\cite{Friedman2001GreedyFA}, including XGBoost ~\cite{Chen2016XGBoostAS}, LightGBM ~\cite{ke2017lightgbm}, and CatBoost ~\cite{conf/nips/ProkhorenkovaGV18}, achieve impressive performance and frequently exceed the performance of deep learning models. Both Gradient Boosting as well as Random Forests generate an ensemble of weak learners composed of decision trees, but they differ in the way those trees are built and combined.

To take advantage of the flexibility of neural networks, various architectures have recently been proposed to improve their performance on tabular data.  Inspired by CatBoost, NODE performs a gradient boosting of oblivious decision trees, which is trained end-to-end using gradient-based optimization ~\cite{popov_neural_2019}. The aim of Net-DNF is to introduce an inductive bias in neural networks corresponding to logical Boolean formulas in disjunctive normal forms ~\cite{katzir2021netdnf}. 
It encourages localized decisions, which involve small subsets of features. TabNet uses a sequential attention mechanism to select a subset of features, which are used at each decision step ~\cite{DBLP:journals/corr/abs-1908-07442}. Hopular is a deep learning architecture in which every layer is composed of continuous modern Hopfield networks ~\cite{schafl2022hopular}. The Hopfield modules allow one to detect various types of dependencies (feature, sample, and target) and have been claimed to outperform concurrent methods on small and medium-sized datasets. The authors of ~\cite{well-tuned} show that the key to boosting the performance of deep learning models is the application of various regularization techniques. They demonstrate that fully connected networks can outperform competitive techniques by applying an extensive search of possible regularizers. The authors of ~\cite{gorishniy_revisiting_2021} introduced modified versions of ResNet and Transformer and showed that the latter outperforms previous neural network models on large datasets. In follow-up papers, the authors worked to transfer the constructed transformer model to other tabular datasets \cite{zhu2023xtab}. Although multiple authors of recent deep learning models often claim to outperform shallow ensemble models, other experimental studies seem to deny these conclusions, showing that typical ensemble methods with careful hyperparameter tuning still presents superior performance~\cite{grinsztajn_why_nodate, shwartz-ziv_tabular_2021}. The authors of \cite{grinsztajn2022tree, mcelfresh2024neural} investigated the situations when deep networks outperform gradient-boosted trees.

In various biological applications, authors try to adapt the architectures created for NLP and CV to the biological domain. In the problem of predicting antimicrobial peptides (AMPs), a language model pre-trained on protein fragments was transferred to classify hemolytic activity \cite{salem2022ampdeep}. The authors of \cite{pandey2023samp} present an image-based deep neural network model to predict AMPs. For this purpose, sequence and structure information is converted into a 3-channel image. In our paper, we go a step further and show that it is possible to transfer ViT pre-trained on images to the case of tabular data. Such an approach reduces the conceptual work on designing the correct architecture for a given problem and minimizes the cost of training the model from scratch.

\section{\our{} model}

In this section, we introduce \our{} -- an adapter network, which allows for a direct transfer of ViT to tabular data. First, we recall the basic idea behind ViT, which is one of the main ingredients of our approach. Next, we discuss possible ways of transferring deep learning models. Finally, we give a detailed description of \our{}.

\subsection{Vision Transformer architecture}

The Vision Transformer (ViT) stands as a monumental shift in how we approach image classification challenges \cite{dosovitskiy_image_2021} and takes inspiration from transformers originally designed for NLP tasks. The fundamental idea is simple, yet powerful. It treats images not as a grid of pixels but as a sequence of smaller, fixed-size patches akin to words in a sentence. Each of these patches is then flattened and projected into a higher-dimensional space, where the sequential processing familiar in NLP tasks is applied.

The architecture comprises several key components, starting with the patch embedding layer. Here, an input image $x \in \R^{H \times W \times C}$ is divided into a sequence of patches $x_1,\ldots,x_n \in \R^{P \times P \times C}$, where $(H,W)$ is the resolution of the image, $C$ is the number of channels, and $P$ is the resolution of the patches. These patches are flattened and transformed into the so-called patch embeddings $t_1,\dots, t_n \in \R^D$ using a trainable linear projection:
\begin{equation} \label{eq:projViT}
f: \R^{P^2 \cdot C} \ni x_i \to t_i \in \R^D.     
\end{equation}
To retain the positional information, which is inherent in image data, position embeddings are added to the patch embeddings, mirroring the process in traditional transformers that deal with text. Additionally, ViT prepends a learnable embedding $\mathrm{CLS}$ to the sequence of embedded patches $T_0 = [\mathrm{CLS}, t_1,\ldots,t_n]$, whose state at the output of the Transformer Encoder serves as the image representation.

Following the embedding layer $f$, the Transformer Encoder is built of multi-head self-attention layers $g_i$, which sequentially transform image representations:
\begin{equation} \label{eq:encViT}
T_i = g_i(T_{i-1}).
\end{equation}
Attention layers $g_i$ allow the model to weigh the importance of different patches in relation to each other, learning global dependencies across the entire image. Unlike conventional convolutional approaches that emphasize local patterns first and more complicated patterns in the deeper layers, the ViT’s attention mechanism inherently allows for the capture of both local and global contextual relationships right from the start, across all layers. 

Finally, the classification head $h$ is attached to the transformed form of the $\mathrm{CLS}$ token to produce the final output. This structure enables ViTs to learn intricate patterns and relationships within the image, leading to their success in various image classification tasks.


\subsection{Transferability}

Basic idea behind transfer learning is that a part of a neural network pre-trained on an upstream task is used for solving a downstream task. In image processing, we typically transfer initial part of the network (a few first layers), which is responsible for extracting basic features of the image. It has been proven that these features are common for various image datasets \cite{yosinski2014transferable}, and, in consequence, there is no need to learn them for each dataset individually. The user supplies this initial part (feature extractor) with custom layers (e.g. classification head) designed to return the response for a downstream task. To use such a network on a downstream task, one can either train only the weights of the newly created output layers, or adjust the whole network (update the weights of the feature extractor and the output layers). The later approach usually works better if there is enough data in the downstream task and we have enough computational budget for performing the training. 

From a practical perspective, it is important to explain what does it mean that a given neural network is transferable between two tasks. To define the transferability, let us consider a neural network, which is composed of two networks $g$ and $h$. First, we pre-train $h_\psi \circ g_\theta$ on task $A$, which results in finding the weights $\psi$ and $\theta$. We usually want to transfer a feature extractor $g$ with pre-trained weights $\theta$ to downstream task $B$. We say that a pre-trained network $g_\theta$ is \emph{transferable} to $B$, if we can find the weights $\phi$ of the network $h'$ such that $h'_\phi \circ g_\theta$ performs at least as well as the network $h' \circ g$ trained from scratch. In other words, reusing the pre-trained weights $\theta$ of $g$ from task $A$ helps in solving the task $B$ using the same architecture. It is not surprising that transferability directly depends on the similarity between tasks $A$ and $B$. Since feature extractors applied to various image data usually find analogical features regardless on the specific dataset, transfer learning in computer vision is possible \cite{yosinski2014transferable}. Here, we investigate the case of transferring ViT encoder from image to tabular data, which is less obvious. 

\subsection{Cross-modal transfer of ViT}

Building upon the foundational principles of transfer learning, we now explore the feasibility of transferring ViT from the image domain to tabular data -- a cross-modal transfer that poses unique challenges.

In the case of ViT, we have patch embedding layer $f$, ViT encoder $g$, and classification head $h$. If we perform transfer inside the image domain it is natural to transfer $g \circ f$ and replace only the classification head $h$. To transfer ViT to tabular data, we cannot directly apply this strategy because the structure of tabular and image data differs. For this reason, we first replace the patch embedding layer $f$ with an adaptation network $\pi$, which is responsible for adjusting tabular input to the form acceptable by ViT encoder. If we now align the distribution of transformed tabular data with the distribution of patch embeddings using adaptation network $\pi$, image and tabular inputs to ViT encoder will become more similar. Forcing similarity between these tabular and patch embeddings will lead to the transferability of the ViT encoder.

According to the definition recalled in the previous subsection, ViT encoder $g_\theta$ with pre-trained weights $\theta$ is \emph{transferable} from image to tabular data, if we can find the weights $\psi$ and $\phi$ such that $h_\psi' \circ g_\theta \circ \pi_\phi$ performs at least as good as $h' \circ g \circ \pi$ trained from scratch on a given tabular task. In this paper, we show that this property holds in most cases for ViT (Table \ref{tab:backbone}).

The introduced adaptation network $\pi$ is used to adjust tabular input $x \in \R^M$ to the form acceptable by the ViT Encoder. It consists of multiple projections $\pi_i: \R^M \to \R^D$, for $i = 1,\ldots,n$. Each projection $\pi_i$ implemented by a simple feed-forward network is responsible for creating a single view of the tabular input $v_i = \pi_i(x)$. These views play a role analogous to the patch embeddings $t_i \in \R^D$ used in ViT. By replacing the patch embedding layer (\ref{eq:projViT}) with the adaptation network $\pi=(\pi_1,\ldots,\pi_n)$, we project tabular data into the patch embedding space, which is the input to the Transformer Encoder (multi-head self-attention layers). 

Next, by supplying tabular views with the $\mathrm{CLS}$ token, we process the sequence $T_0 = [\mathrm{CLS}, v_1, \ldots,v_n]$ by the ViT encoder (\ref{eq:encViT}) pre-trained on image data. Finally, we replace the original ViT classification head $h$ by the network $h'$ responsible for classifying tabular inputs. While the introduction of the adaptation layer $\pi$ is a unique feature of the cross-modal transfer, the modification of the classification head is a common step in transfer learning and, particularly, in fine-tuning ViT.

In a typical strategy of training \our{}, the parameters of the ViT encoder $g$ are frozen and do not change during training. We only modify (train) the weights of the adaptation network $\pi$ and the classification head $h'$. Due to the small number of trainable parameters compared to the complexity of the whole \our{} model, we can use the benefits of a large model trained at relatively low cost. In particular, this allows us to use \our{} on small tabular datasets. Alternatively, we can fine-tune the whole model and adjust the parameters of the ViT encoder as well. In \Cref{tab:backbone}, we show that this approach can often increase the final score.

Our findings shed a new light on the area of transfer learning. First, we demonstrate that transfer learning goes beyond using pre-trained feature extractor and can be applied to middle layers of the network. Second, we show that it is possible to effectively perform a cross-modal transfer from image to tabular data. In cross-modal transfer, we use a large-scale model with pre-trained dependencies, but at the same time, we avoid the computationally expensive process of training it from the ground up. This is especially profitable in training of deep models on small tabular data containing less than 1k samples, which are ubiquitous in the tabular domain.

\section{Experiments}

This section presents the experimental evaluation of \our{}. We start by comparing \our{} with state-of-the-art shallow and deep methods in tabular data classification. Next, we investigate the application of \our{} in the case of an extremely small number of training data. Finally, we investigate various aspects of \our{} model, such as type of the ViT encoder, depth of the projection and output networks, fine-tuning techniques, and using only the part of the ViT encoder in the architecture transfer.

\subsection{Tabular Data Classification}

First, we benchmark \our{} against well-established shallow methods and recent deep learning models on publicly available examples of tabular data in the classification tasks. To take the advantage of our transfer learning approach, we intentionally focus on small datasets with less than 1k samples, in which \our{} performs best. At the end of this subsection, we evaluate \our{} in the few shot scenario.

\paragraph{Experimental setup}

We consider small datasets retrieved from the UCI repository, which are summarized in Table~1 in Appendix A\footnote{Available at \url{https://github.com/wwydmanski/VisTabNet/blob/main/Appendix.pdf}}. 
Small datasets are the most challenging case for deep learning methods, but thanks to applying transfer learning principle, \our{} is capable of reducing the overfitting issue.

\our{} is compared to the following methods: (i) {\bf RF}: Random Forests~\cite{Breiman2001RandomF}, (ii) {\bf GB} :Gradient Boosting~\cite{Friedman2001GreedyFA}, (iii) {\bf XGBoost}~\cite{Chen2016XGBoostAS}, (iv) {\bf LightGBM}~\cite{ke2017lightgbm}, (v) {\bf ResNet}~\cite{gorishniy_revisiting_2021}, (vi) {\bf FT}: Feature Transformer~\cite{gorishniy_revisiting_2021}, (vii) {\bf NODE}: Neural Oblivious Decision Ensembles)~\cite{popov_neural_2019}.
These methods were selected due to their popularity and proven effectiveness in tabular data classification tasks, serving as a comprehensive baseline for measuring VisTabNet's performance~\cite{grinsztajn2022tree, mcelfresh2024neural}.

\begin{table*}[!htb]
  \centering
  \caption{Benchmark of VisTabNet against other commonly used algorithms using Matthews Correlation Coefficient (the higher the better) accompanied with the corresponding standard deviation. \our{} obtains the highest mean MCC score and the best mean rank.
  }
  \small
  \label{tab:benchmark}
    \begin{tabular}{lcccccccc}
      \toprule
   Dataset & VisTabNet & RF & XGBoost & GB & LightGBM & ResNet & FT & NODE \\ 
    \midrule
    Blood transf.       & 31.3 $\pm$ 7 & 22.0 $\pm$ 3                  & 30.4 $\pm$ 4 & 30.4 $\pm$ 4 & 30.4 $\pm$ 4 & \textbf{\uline{45.3 $\pm$ 6}} & 41.6 $\pm$ 6 & 28.5 $\pm$ 6 \\
    Wisconsin           & \textbf{\uline{65.3 $\pm$ 5}} & 33.0 $\pm$ 3                  & 30.6 $\pm$ 4 & 30.6 $\pm$ 4 & 30.6 $\pm$ 4 & 30.6 $\pm$ 5 & 31.7 $\pm$ 5 & 30.6 $\pm$ 2 \\
    Breast Cancer       & 91.1 $\pm$ 4 & 88.4 $\pm$ 2   & 80.7 $\pm$ 3                  & 87.0 $\pm$ 3 & 89.6 $\pm$ 3 & \textbf{\uline{97.3 $\pm$ 6}} & 94.6 $\pm$ 4 & 92.5 $\pm$ 18 \\
    Connectionist       & \textbf{\uline{84.6 $\pm$ 5}} & 69.0 $\pm$ 3                  & 76.2 $\pm$ 4 & 74.6 $\pm$ 4 & 63.6 $\pm$ 4 & 64.5 $\pm$ 7 & 37.7 $\pm$ 5 & 76.3 $\pm$ 4 \\
    Congr. Voting       & 91.5 $\pm$ 4                  & 93.7 $\pm$ 2         & 91.7 $\pm$ 3 & \textbf{\uline{95.7 $\pm$ 3}} & 90.3 $\pm$ 3 & 73.9 $\pm$ 6 & 79.9 $\pm$ 4 & 89.7 $\pm$ 2 \\
    Credit Approval     & 67.5 $\pm$ 1      & 74.1 $\pm$ 3                  & 74.3 $\pm$ 4 & 71.1 $\pm$ 4 & 74.1 $\pm$ 4 & 65.9 $\pm$ 7 & 74.9 $\pm$ 5 & \textbf{\uline{79.9 $\pm$ 5}} \\
    Cylinder bands      & \textbf{\uline{45.0 $\pm$ 4}} & 44.3 $\pm$ 3                  & 33.4 $\pm$ 4 & 33.4 $\pm$ 4 & 42.7 $\pm$ 4 & 43.7 $\pm$ 6 & 39.7 $\pm$ 6 & 44.4 $\pm$ 8 \\
    Dermatology         & 95.3 $\pm$ 1                  & \textbf{\uline{96.5 $\pm$ 2}} & 95.3 $\pm$ 3 & 93.1 $\pm$ 3 & 95.2 $\pm$ 3 & 84.9 $\pm$ 6 & 92.3 $\pm$ 4 & 91.1 $\pm$ 3 \\
    Ecoli               & 72.1 $\pm$ 5                  & 76.2 $\pm$ 3                  & 70.3 $\pm$ 4 & 68.3 $\pm$ 4 & 70.2 $\pm$ 4 & 87.1 $\pm$ 7 & 89.6 $\pm$ 5 & \textbf{\uline{90.1 $\pm$ 4}} \\
    Glass               & 93.9 $\pm$ 4                  & 93.8 $\pm$ 2                  & 95.9 $\pm$ 3 & 95.9 $\pm$ 3 & 95.9 $\pm$ 3 & 64.6 $\pm$ 6 & 58.0 $\pm$ 4 & \textbf{\uline{100.0 $\pm$ 0}} \\
    Haberman            & \textbf{\uline{50.2 $\pm$ 6}} & 24.6 $\pm$ 3                  & 27.8 $\pm$ 4 & 25.8 $\pm$ 4 & 30.4 $\pm$ 4 & 27.1 $\pm$ 7 & 40.1 $\pm$ 6 & 31.8 $\pm$ 12 \\
    Horse Colic         & 50.6 $\pm$ 5                  & \textbf{\uline{75.4 $\pm$ 3}} & 75.1 $\pm$ 4 & 75.1 $\pm$ 4 & 58.1 $\pm$ 4 & 43.1 $\pm$ 8 & 43.1 $\pm$ 5 & 57.4 $\pm$ 3 \\
    Ionosphere          & 87.7 $\pm$ 4         & 83.4 $\pm$ 2                  & 79.4 $\pm$ 3 & 77.3 $\pm$ 3 & 69.6 $\pm$ 3 & 87.0 $\pm$ 6 & \textbf{\uline{95.7 $\pm$ 4}} & 77.6 $\pm$ 19 \\
    Libras              & \textbf{\uline{84.4 $\pm$ 3}} & 70.7 $\pm$ 3                  & 66.9 $\pm$ 4 & 63.0 $\pm$ 4 & 70.7 $\pm$ 4 & 77.5 $\pm$ 7 & 59.7 $\pm$ 5 & 59.7 $\pm$ 5 \\
    Lymphography        & 70.7 $\pm$ 5         & 66.8 $\pm$ 3                  & 47.7 $\pm$ 4 & 66.8 $\pm$ 4 & 41.4 $\pm$ 4 & 58.9 $\pm$ 7 & 42.7 $\pm$ 5 & \textbf{\uline{72.1 $\pm$ 19}} \\
    Mammographic        & 60.1 $\pm$ 5                  & 68.6 $\pm$ 3                  & 72.6 $\pm$ 4         & 69.3 $\pm$ 4 & 70.9 $\pm$ 4 & 72.5 $\pm$ 6 & \textbf{\uline{73.8 $\pm$ 5}} & 64.7 $\pm$ 12 \\
    Primary Tumor       & \textbf{\uline{40.1 $\pm$ 6}} & 30.6 $\pm$ 3                  & 34.6 $\pm$ 4 & 36.0 $\pm$ 4 & 35.2 $\pm$ 4 & 32.5 $\pm$ 7 & 39.1 $\pm$ 6 & 39.6 $\pm$ 9 \\
    Sonar               & 63.0 $\pm$ 5                  & 63.0 $\pm$ 3                  & 62.2 $\pm$ 4 & 63.0 $\pm$ 4 & \textbf{\uline{68.8 $\pm$ 4}} & 36.0 $\pm$ 7 & 78.0 $\pm$ 5 & 60.1 $\pm$ 4 \\
    Statlog Australian  & 70.9 $\pm$ 5                  & 71.8 $\pm$ 3                  & 72.0 $\pm$ 4 & 73.5 $\pm$ 4 & 71.3 $\pm$ 4 & 67.5 $\pm$ 7 & \textbf{\uline{74.9 $\pm$ 5}} & 60.8 $\pm$ 6 \\
    Statlog German      & 29.3 $\pm$ 6                  & \textbf{\uline{43.1 $\pm$ 3}} & 39.2 $\pm$ 4 & 39.2 $\pm$ 4 & 39.2 $\pm$ 4 & 41.0 $\pm$ 7 & 37.3 $\pm$ 6 & 42.5 $\pm$ 14 \\
    Statlog Heart       & 40.3 $\pm$ 5                  & 55.4 $\pm$ 3 & 58.3 $\pm$ 4   & 52.4 $\pm$ 4 & 52.4 $\pm$ 4 & 62.3 $\pm$ 7 & \textbf{\uline{78.0 $\pm$ 5}} & 43.7 $\pm$ 3 \\
    Vertebral           & 70.6 $\pm$ 5                  & \textbf{\uline{74.6 $\pm$ 3}} & 73.5 $\pm$ 4 & 58.7 $\pm$ 4 & 71.9 $\pm$ 4 & 67.6 $\pm$ 7 & 68.9 $\pm$ 5 & 65.7 $\pm$ 4 \\
    Zoo                 & 94.3 $\pm$ 2                  & 94.6 $\pm$ 2                  & 94.6 $\pm$ 3 & \textbf{\uline{100.0 $\pm$ 0}} & 94.6 $\pm$ 1 & 81.0 $\pm$ 6 & 81.0 $\pm$ 4 & 94.6 $\pm$ 6 \\
    \midrule
    Mean & \textbf{\uline{67.43}} & 65.81 & 64.47 & 64.36 & 63.35 & 61.38 & 63.14 & 64.93 \\
    Mean rank & \textbf{\uline{3.93}} & 4.04 & 4.39 & 4.91 & 4.87 & 5.17 & 4.24 & 4.43 \\
    \bottomrule
    \end{tabular}%
\end{table*}

We apply double cross-validation procedure. The hyperparameters are selected using train-validation splits, while the models' performance is reported on train-test split.
For each dataset, we perform careful hyperparameter optimization using PyHopper library, executing 50 optimization steps with four running in parallel and a seeding ratio of 0.5. The best hyperparameters are the ones that perform best on the validation set, so the test set is never used for tuning. Each method uses identical train-validation-test splits. To avoid random effects, the experiments are repeated three times on different splits. Addressing the potential issue of class imbalance, we employ the RandomOverSampler to resample the training dataset.

As an evaluation metric, we employ Matthews Correlation Coefficient (MCC)~\cite{matthews_comparison_1975}, which is known to be robust to imbalance classification problems~\cite{chicco_matthews_2021}.
It calculates the correlation coefficient between the observed and predicted classifications, producing a value that ranges from -1 to 1. A coefficient of 1 signifies a perfect prediction, 0 is no better than random guessing, and -1 indicates total disagreement between prediction and observation. 

\paragraph{Results}

\our{} achieves the highest average MCC score and obtains the best rank, see \Cref{tab:benchmark}. Its mean score is 2.5 percentage points higher than the second-best deep model (NODE) and 1.62 percentage points higher than the best shallow model (RF). It demonstrates that the cross-modal transfer applied by \our{} is more effective than training deep networks from scratch, especially in the context of small datasets. While the competitive transformer model (FT) obtains relatively good rank, it failed to succeed on multiple datasets, which resulted in worse mean MCC score. The results also confirm that shallow methods represent strong baselines, which are difficult to outperform by advanced deep models. Moreover, comparing the standard deviations show that the performance of \our{} is more stable than competitive deep models.



\paragraph{Few-shot transfer learning}

Transfer learning is extremely efficient in the case of small sample problems. In this part, we consider an extreme case, where only a few examples of each class are available in a downstream task (from 1 to 10 examples per class), which is analogous to $N$-shot scenario. We restrict our attention to 5 datasets (Credit Approval, Cylinder Bands, Dermatology, Libras, Zoo) and shallow methods, which are not so prone to overfitting as deep models.

\begin{figure}[!htb]
  \subfigure[Mean of MCC scores (the higher the better)]{\includegraphics[width=0.49\textwidth]{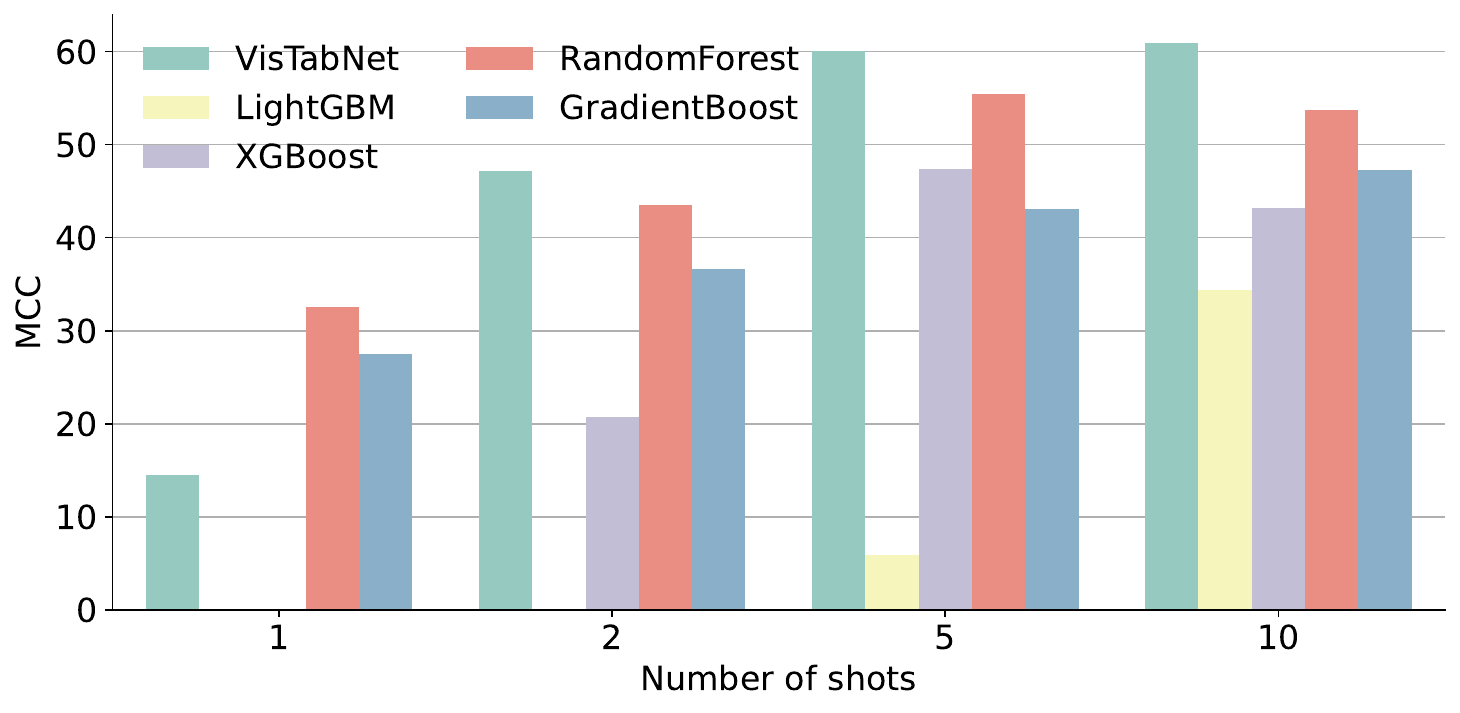}}
  \subfigure[Boxplot of ranking (the lower the better)]{\includegraphics[width=0.49\textwidth]{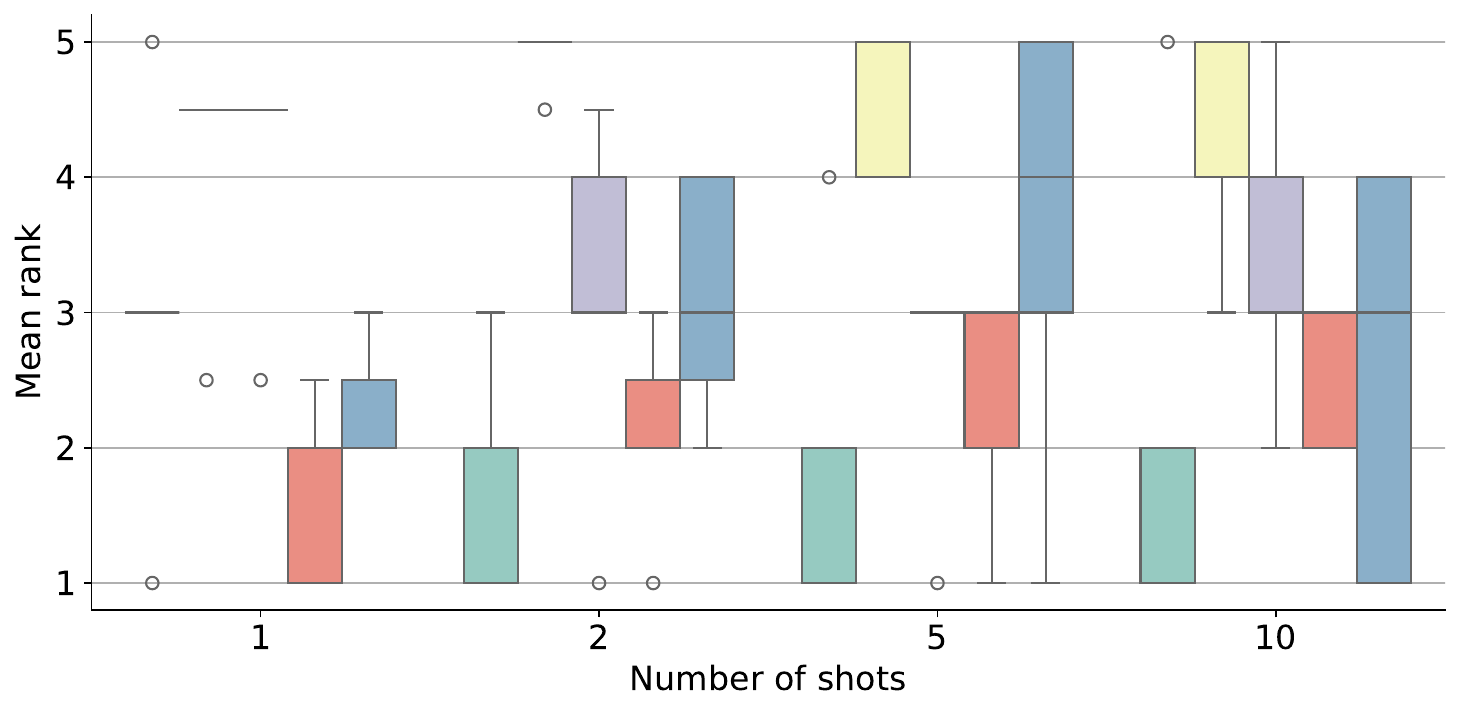}}
  \caption{Average performance on 5 datasets (Credit Approval, Cylinder Bands, Dermatology, Libras, ZOO) in $N$-shot setting with $N=1, 2, 5, 10$. VisTabNet achieves significantly better scores in the few-shot setting, consistently outperforming other training methods between 2 and 10 shot.} \label{fig:few}
\end{figure}

The results presented in Figure \ref{fig:few} show that VisTabNet outperforms the rest of the approaches when more than 2 examples per class were available. While RF and GB return better results for 1-shot case, they are not able to use as much information from more examples as \our{}. It confirms superior transfer learning capabilities of \our{}.

\subsection{Analysis of \our{} components} \label{sec:analysis}

In this part, we analyze the main building blocks of \our{}. We investigate the influence of finetuning techniques, the selection of transformer encoder, depth of the adaptation and classification networks as well as reduction of layer in the ViT encoder. This analysis was also conducted on 5 additional datasets: Credit Approval, Cylinder Bands, Dermatology, Libras, and Zoo.

\paragraph{Backbone selection}

\our{} can be instantiated with various ViT architectures, e.g. ViT Base, or ViT Large. There appears a question of how the selection of ViT backbone influences the final performance of the model. Second question concerns the selection of the optimization procedure. We can either (i) train only the adaptation and output networks as it was done in our main benchmark, or (ii) fine-tune the ViT encoder after initial training of the adaptation and output networks, or (iii) train all components of \our{} at once (including ViT encoder). Finally, we can ask what is a benefit of applying pre-trained ViT encoder. For this purpose, we compare \our{} with dense neural network composed of adapter and classification networks (without ViT encoder). 

\begin{table*}[!htb]
\centering
  \caption{Influence of the training strategy (fine-tuned, fully-trained) of \our{} and the selection of ViT encoder (base vs. large). We additionally show that removing ViT encoder from the \our{} architecture significantly decreases its performance.}
  \label{tab:backbone}
  \small
  \begin{tabular}{lccccc}
    \toprule
    Dataset & VisTabNet (B) & VisTabNet (B) & VisTabNet (B) & VisTabNet (L) & No ViT encoder\\
     &  & fine-tuned & fully-trained & & \\
    \midrule
    Dermatology & 0.930  & 0.920 & 0.930 &  \textbf{\uline{0.957}} &0.842\\
    Libras & 0.843 & \textbf{\uline{0.853}} & 0.812 & 0.812 & 0.701\\
    ZOO & \textbf{\uline{0.946}}  &  0.891 & 0.838 & 0.838 & 0.733\\
    Cylinder Bands & \textbf{\uline{0.426}}  & \textbf{\uline{0.426}} & 0.418 & 0.413  & 0.407\\
    Credit approval & 0.651 & \textbf{\uline{0.665}} & 0.639  & 0.626 & 0.580\\ \midrule
    Mean rank & \textbf{\uline{1.8}} & 1.9 & 3.1 & 3.2 & 5\\
    \bottomrule
  \end{tabular}
\end{table*}

Our findings presented in Table \ref{tab:backbone} indicate that VisTabNet (B) generally outperforms VisTabNet (L). This suggests that increasing the model size does not necessarily translate to better performance, particularly in the context of small datasets. 

Training all parameters of \our{} at once (fully-trained case) significantly deteriorates the performance of \our{}. The effect of fine-tuning ViT encoder after training adapter and classification networks is moderate: in 2 cases finetuning improves the results, on 1 dataset is has no effect, while in two remaining cases it deteriorates model's performance. Decrease in accuracy could be attributed to too high learning rate in the finetuning stage. Additional experiments with manually selected learnig rate led to stabilization of the results. It suggests that fine-tuning could be considered as an additional hyperparameter in \our{}. 

We also highlight a significant insight: adapting a ViT architecture as the backbone for \our{} has significant influence on the performance (last column). This advantage is observed regardless of the specific backbone selection, underscoring the efficacy of leveraging pre-existing architectures designed for different tasks. 
The success of this strategy reinforces the value of cross-modal learning and the adaptability of transformer architectures, setting a promising direction for future research in machine learning methodologies.

\paragraph{Transformer architectures}

Instead of transferring ViT Encoder, we can use alternative transformer models. In this experiment, we investigate the transfer of BERT architecture pre-trained on NLP task.

\begin{table*}[!htb]
\centering
\caption{Performance of \our{} with 3 pre-trained BERT architectures compared to its standard variant based on ViT.}
\small
\begin{tabular}{lcccc}
\toprule
Dataset   & BERT Tiny       & BERT Mini       & BERT Small    & VisTabNet \\ 
  &     &        & & ViT base \\ \midrule
Cylinder bands  & 43.7 $\pm$ 1   & 44 $\pm$ 5     & 38 $\pm$ 6            & \textbf{\uline{45 $\pm$ 4}} \\ 
Credit Approval & 66.2 $\pm$ 1   & 66 $\pm$ 2     & 66 $\pm$ 1         & \textbf{\uline{67.5 $\pm$ 1}} \\
Dermatology     & 92.9 $\pm$ 4   & \textbf{\uline{96 $\pm$ 1}}     & 95 $\pm$ 0 &  95.3 $\pm$ 1  \\
Libras          & 82.7 $\pm$ 2   & 79 $\pm$ 1     & 77 $\pm$ 2    &  \textbf{\uline{84.4 $\pm$ 3}}  \\
ZOO             & 92.9 $\pm$ 2   & \textbf{\uline{94.6 $\pm$ 0}}   & 92.9 $\pm$ 3     & 94.3 $\pm$ 2 \\  \midrule
Mean rank & 2.9 & 2.1 & 3.6  &\textbf{\uline{1.4}} \\ \bottomrule
\label{tab:architectures}
\end{tabular}
\end{table*}

The results presented in \Cref{tab:architectures} demonstrate that \our{} with a pre-trained ViT model achieves superior performance in most cases, obtaining the lowest mean rank of 1.4. This model exhibits particularly high efficacy on the Cylinder bands (45 ± 4), Credit Approval (67.5 ± 1), and Libras (84.4 ± 3) datasets, where it attains the highest scores.
Among the BERT models, BERT Mini demonstrates the best overall performance with a mean rank of 2.1, but its results are less consistent compared to VisTabNet (pre-trained ViT). Nevertheless, these findings suggest that it is worth to investigate alternative cross-modal transfer since the BERT encoder provide overall promising results.


\paragraph{Depth of adaptation and classification networks}

\our{} is paramterized by the adaptation and classification networks. We investigate how the number of layers in these networks affects the final performance.

\begin{figure}[!htb]
  \centering
  \includegraphics[width=\linewidth]{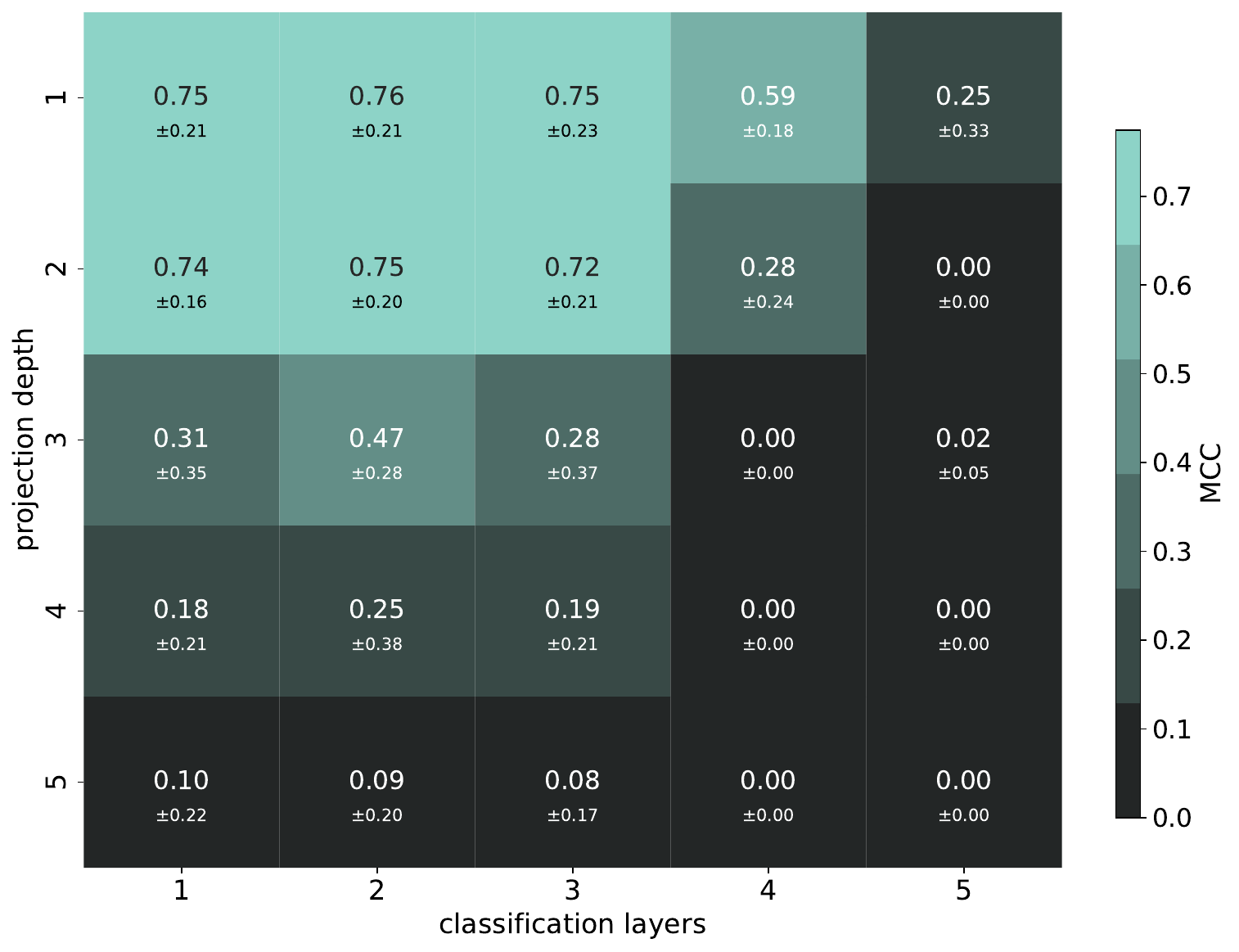}
  \caption{Influence of the depth of adaptation and classification networks on the \our{} performance using 5 datasets (ZOO, Dermatology, Credit Approval, Cylinder Bands, Libras).} \label{tab:focal}
\end{figure}

The results presented in \Cref{tab:focal} indicates that using small networks generally works best. \our{} confirms high performance for 1-2 adaptation layers and 1-3 output layers. For larger number of layers, the accuracy of \our{} significantly drops, which again can be attributed to small-sample problems investigated in this paper. 

\paragraph{Reduction of ViT Encoder}

In the base version, \our{} transfers the entire ViT encoder. In the cross-modal transfer, we can use an arbitrary part of the pre-trained network and we are not forced to use the entire encoder. In this experiment, we examine what part of the ViT encoder has to be transferred to obtain the best performance. As we decreased the number of layers traversed, we generally reduce the computing time.

\begin{figure}[!ht]
    \centering
    \includegraphics[width=\linewidth]{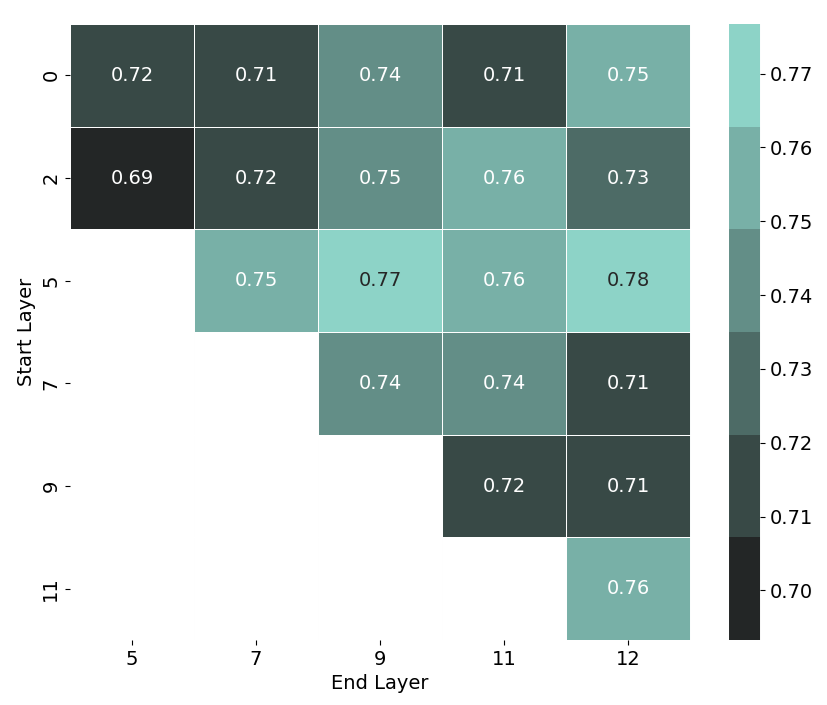}
    \caption{Average performance of \our{} when selected layers were removed from the ViT encoder using 5 datasets (ZOO, Dermatology, Credit Approval, Cylinder Bands, Libras).}
    \label{fig:projection hitmap}
\end{figure}

The heatmap presented in Figure \ref{fig:projection hitmap} shows the average MCC scores obtained by transferring the part of the ViT encoder ranging from the "start layer" to the "end layer". The standard variant of \our{} visualized in the top right corner (start = 0, end = 12) results in the MCC value of 0.75, which is 3 percentage points lower than the best score obtained going from the 5th to 12th layer. It can be observed that projecting data onto the 5th layer gives generally very good results. While initial layers of the ViT encoder are pre-trained to process linearly transformed image patches, later layers work on more abstract representations, which may better suit to tabular data. In consequence, if we start from the first layer of the ViT encoder, we need to apply many transformation (ending in the 12th layer) to get meaningful representation. Starting with more general transformations defined by the 5th layer, we do not need to apply so deep networks, which provide good balance between performance (measured by the MCC) and computation time (network reduction led to a 2-3 times reduction in computational time). Detailed results on individual datasets can be found in Figure 1 of Appendix B.

\paragraph{Conclusion of the analysis} While the basic version of \our{} gives very competitive results to the state-of-the-art methods, detailed analysis presented in this section suggests that the performance of \our{} can be further improved. First of all, one should investigate reducing the complexity of ViT encoder by eliminating its initial layers. Second, slight fine-tuning of the entire \our{} model after training adaptation and classification network often leads to the slight improvements of the results. Finally, the proposed cross-modal transfer can also be applied to NLP models, which should be investigated in more details in the future works.

\section{Conclusion}

In this paper, we introduced a cross-modal transfer, which allows for reusing a neural network pre-trained on images to process tabular data. This idea was realized on the ViT architecture, in which we replaced patch embedding network with an adaptation layer. By forcing the similarity between transformed tabular inputs and the embeddings of image patches, we obtained transferability of ViT encoder with a minimal conceptual and computational cost. Our approach demonstrates that transfer learning goes beyond reusing feature extractor in computer vision, and can be applied to middle layers of neural networks as well as is feasible in cross-modal setting. As a future work, we leave the question whether a cross-modal transfer can be applied to network architectures different from transformers. Finding positive answers to this problem can open up new avenues in transfer learning.

\section*{Acknowledgements}

This research has been supported by the flagship project entitled “Artificial Intelligence Computing Center Core Facility” from the Priority Research Area DigiWorld under the Strategic Programme Excellence Initiative at Jagiellonian University. This research was supported by the National Science Centre (Poland), grant no. 2023/50/E/ST6/00169. We gratefully acknowledge Polish high-performance computing infrastructure PLGrid (HPC Center: ACK Cyfronet AGH) for providing computer facilities and support within computational grant no. PLG/2023/016610. The work of Witold Wydmański is supported by the Ministry of Science grant no. PN/01/0195/2022. We would like to thank \L{}ukasz Struski for invaluable discussions and his help at the initial stage of this work.

For the purpose of Open Access, the author has applied a CC-BY public copyright license to any Author Accepted Manuscript (AAM) version arising from this submission.

\bibliographystyle{siam}
\bibliography{sample-base}

\appendix

\section{Experimental setup} \label{appendix}

To aid in reproducing the results, we present technical details regarding our experiments. 

Initially, we divided the dataset into training and testing parts, allocating three-quarters of the data for training and the remaining quarter for testing. 
Subsequently, both the training and testing parts were preprocessed based on the characteristics observed in the training set, ensuring that the models were trained on data representative of the real-world scenarios they would encounter.

To further refine the training process, the training dataset was then split again, this time into training and validation datasets with proportions of four fifths and one fifth, respectively. This resulted in final proportions of the train, valid, and test sets being 12/20, 3/20, and 5/20 of the entire dataset. This split was instrumental in tuning the models and preventing overfitting.

Upon completion of hyperparameter optimization, the training and validation datasets were merged into a single \textit{full\_train} dataset. The models then underwent final training on this \textit{full\_train} dataset, utilizing the hyperparameters identified as optimal in the previous step. This comprehensive training regime, culminating in testing on the separate test split, was designed to mitigate any risk of cross-contamination in the results, ensuring the integrity and reliability of our findings.

Hyperparameter optimization was performed using the PyHopper library, executing 50 optimization steps with four running in parallel and a seeding ratio of 0.5. This optimization was carried out on the train/validation splits, allowing us to fine-tune the models for optimal performance. We used the following ranges of hyperparameters for each method:

\paragraph{LightGBM}
\begin{verbatim}
num_leaves = choice(2, 4, 8, 16, 32, 64),
max_depth = choice(-1, 2, 4, 8, 16, 32, 64),
learning_rate =float(0.001, 0.1, log=True),
n_estimators = choice(10, 50, 100, 200, 500, 
1000)
\end{verbatim}

\paragraph{XGBoost}
\begin{verbatim}
n_estimators = int(50, 1000, multiple_of=50, 
init=50),
max_depth = choice(2, 3, 5, 10, 15),
learning_rate = float(1e-5,1e-1, log=True),
min_child_weight = choice(1, 2, 4, 8, 16, 32),
gamma = choice(0, 0.001, 0.1, 1)
\end{verbatim}

\paragraph{Random Forest}
\begin{verbatim}
n_estimators = int(50, 3000, multiple_of=50),
max_features = choice(None, 'sqrt', 0.2, 0.3, 
0.5, 0.7),
criterion = choice('gini', 'entropy'),
max_depth = choice(None, 2, 4, 8, 16)
\end{verbatim}

\paragraph{Gradient Boosting}
\begin{verbatim}
n_estimators = int(50, 3000, multiple_of=50, 
init=50),
max_depth = choice(2, 3, 5, 10, 15),
learning_rate = float(1e-5,1e-1, log=True)
\end{verbatim}

\paragraph{NODE}
\begin{verbatim}
layer_dim = int(64, 1024, power_of=2),
num_layers = int(1, 5),
depth = int(2, 7)
\end{verbatim}


\paragraph{VisTabNet}
\begin{verbatim}
LR = float(1e-5, 1e-3, "0.1g"),
PROJ_LR = float(1e-5, 1e-3, "0.1g"),
EPOCHS = int(10, 100, multiple_of=10),
PROJECTIONS = choice(8, 16, 32, 64, 128),
PROJ_DEPTH = choice(1, 2, 3, 4)
\end{verbatim}

\paragraph{ResNet}
\begin{verbatim}
EPOCHS = choice(10, 30, 50, 100, 150),
PATIENCE = choice(2, 5, 10, 16, 24, 37)
\end{verbatim}

\paragraph{FT}
\begin{verbatim}
N_BLOCK: choice(1,2,3,4,5,6),
D_BLOCK: choice(96, 128, 192, 256, 320, 384),
ATTENTION_DROPOUT: choice(0.1, 0.15, 0.2,
0.25, 0.3, 0.35),
FFN_DROPOUT: choice(0.0, 0.05, 0.1, 0.15, 0.2, 
0.25),
EPOCHS = choice(50, 100, 150),
PATIENCE = choice(2, 10, 16, 37)
\end{verbatim}

The experiments were conducted on 20 datasets, which are summarized in Table \ref{tab:datasets}.

\begin{table}[!htb]
  \centering
  \caption{Summary of the datasets. \label{tab:datasets}}
  \begin{tabular}{lcccc}
    \toprule
    Dataset & Size & Continuous & Categorical & Classes \\
     &  & Attributes & Attributes & \\
    \midrule
    Blood Trans.  & 748 & 4 & 1 & 2 \\
    BC Wisconsin  & 569 & 30 & 0 & 2 \\
    Breast Cancer  & 286 & 0 & 9 & 2 \\
    Connectionist  & 208 & 60 & 0 & 2 \\
    Congr. Voting  & 435 & 0 & 16 & 2 \\
    Credit Approval  & 690 & 6 & 9 & 2 \\
    Cylinder Bands  & 512 & 20 & 19 & 2 \\
    Dermatology  & 366 & 34 & 0 & 6 \\
    Ecoli &  336 & 5 & 0 & 8 \\
    Glass  & 214 & 9 & 0 & 6 \\
    Haberman  & 306 & 3 & 0 & 2 \\
    Horse Colic  & 368 & 8 & 19 & 2 \\
    Ionosphere  & 351 & 34 & 0 & 2 \\
    Libras & 360 & 90 & 0 & 15 \\
    Lymphography & 148 & 18 & 0 & 4 \\
    Mammographic & 961 & 1 & 5 & 2 \\
    Primary Tumor & 330 & 0 & 17 & 21 \\
    Sonar & 208 & 60 & 0 & 2 \\
    Statlog Australian & 690 & 5 & 9 & 2 \\
    Statlog German & 1000 & 23 & 0 & 2 \\
    Statlog Heart & 270 & 6 & 7 & 2 \\
    Vertebral & 310 & 6 & 0 & 2 \\
    Zoo & 101 & 16 & 0 & 7 \\
    \bottomrule
  \end{tabular}
\end{table}

\section{Detailed results}

\begin{figure*}[!ht]
    \centering
    \subfigure[ZOO]{\includegraphics[width=0.45\linewidth]{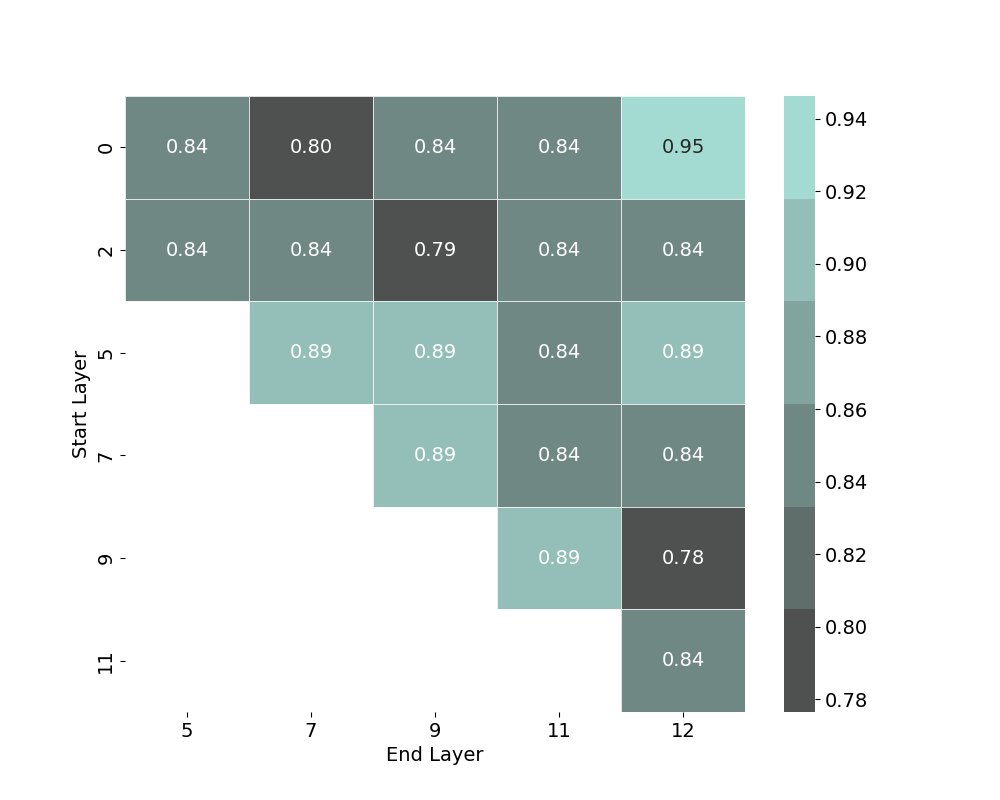}} 
    \subfigure[Dermatology]{\includegraphics[width=0.45\linewidth]{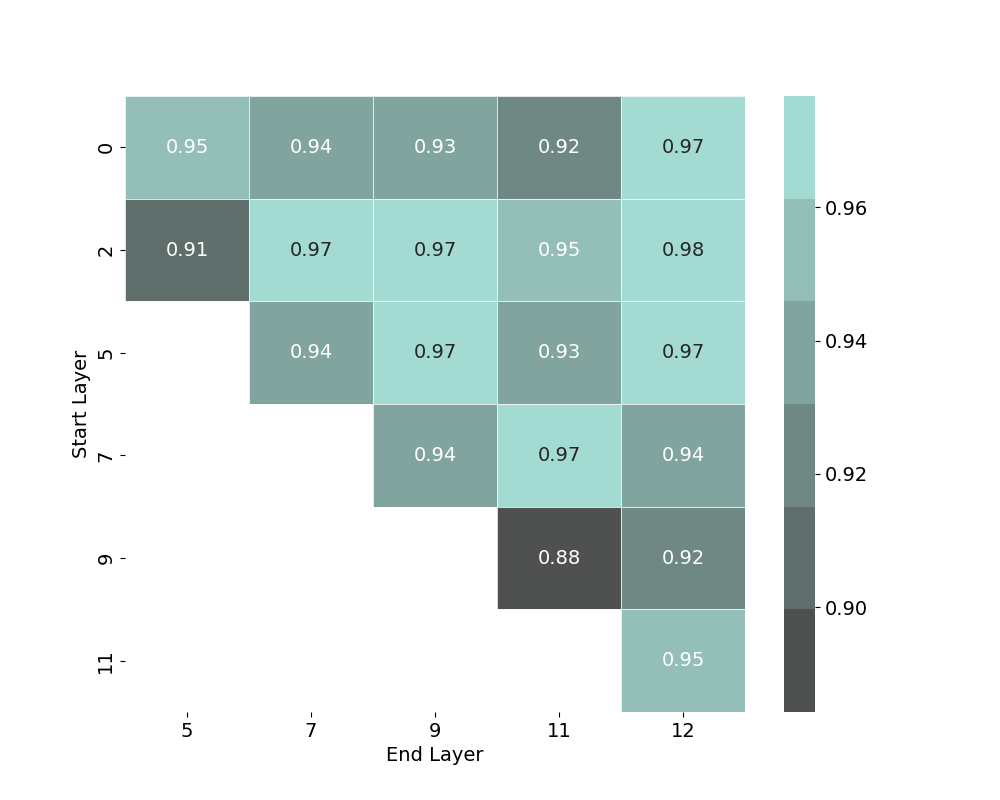}} 
    \subfigure[Credit Approval]{\includegraphics[width=0.45\linewidth]{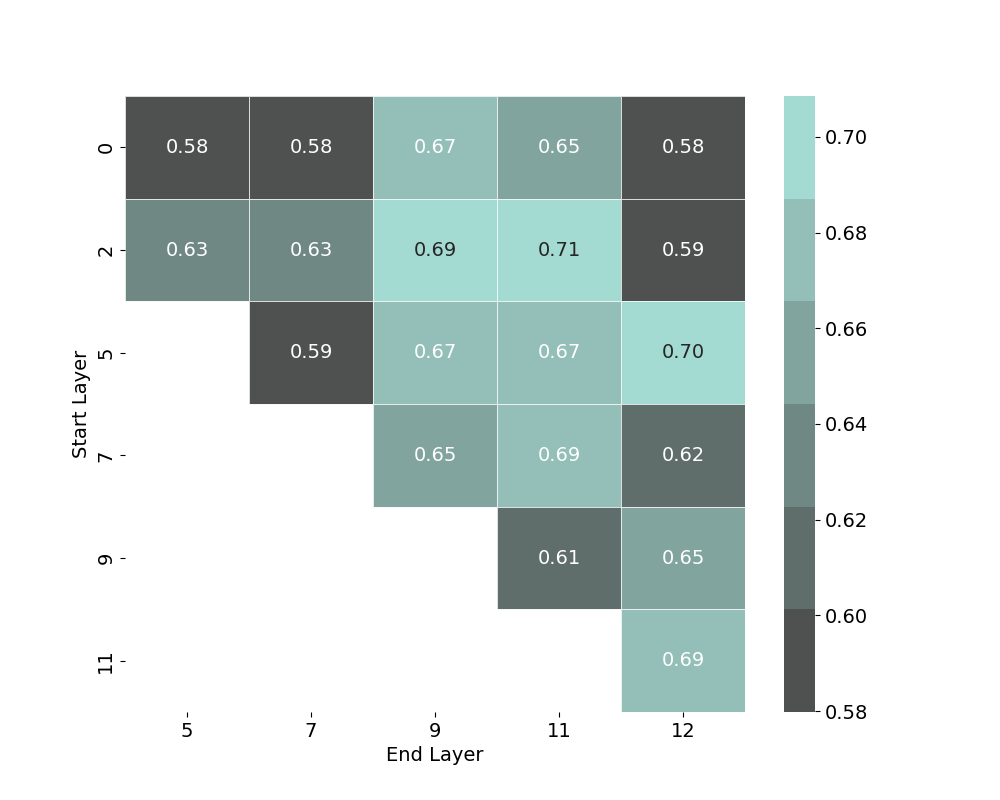}} 
    \subfigure[Libras]{\includegraphics[width=0.45\linewidth]{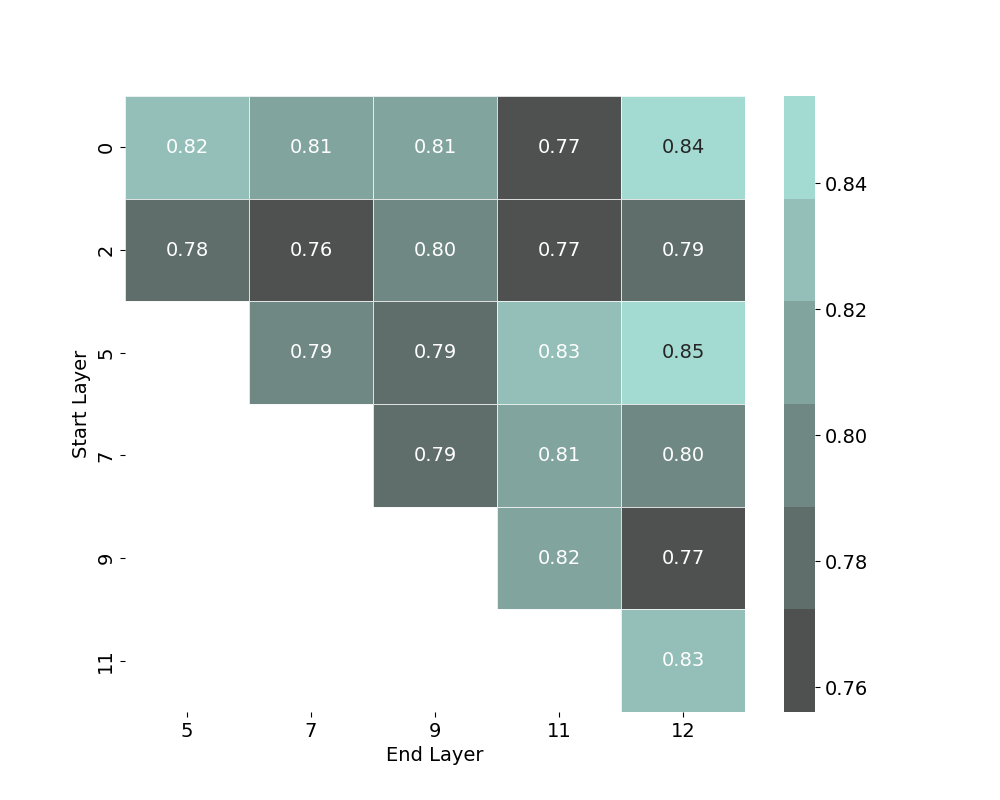}} 
    \subfigure[Cylinder Bands]{\includegraphics[width=0.45\linewidth]{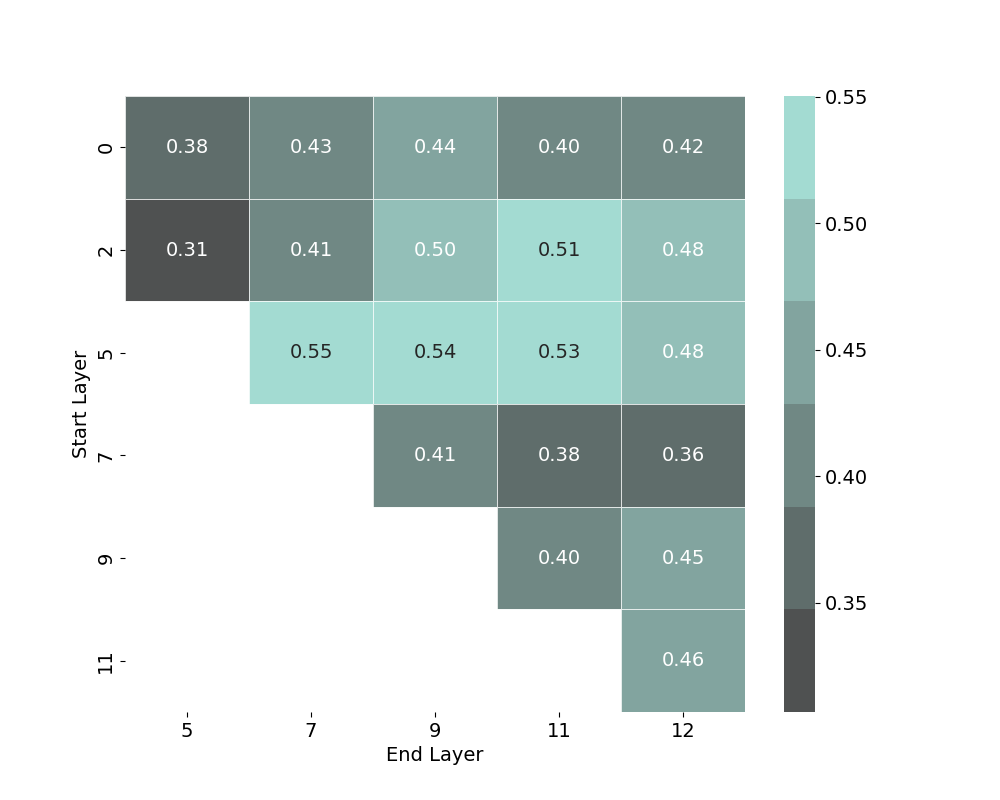}} 
    \caption{Performance of \our{} when selected layers were removed from the ViT encoder.}
    \label{fig:reduction}
\end{figure*}

Figure \ref{fig:reduction} presents the MCC score of \our{} when only the part of the ViT encoder was transferred to \our{} architecture. Other layers were completely removed.  

\begin{figure*}[!ht]
    \centering
    \subfigure[ZOO]{\includegraphics[width=0.3\linewidth]{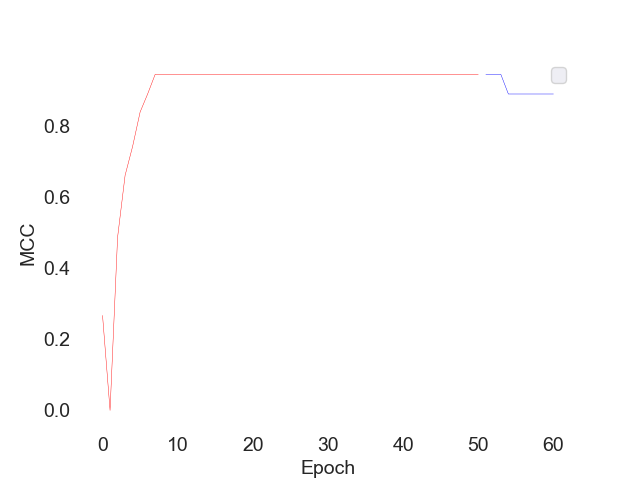}} 
    \subfigure[Dermatology]{\includegraphics[width=0.3\linewidth]{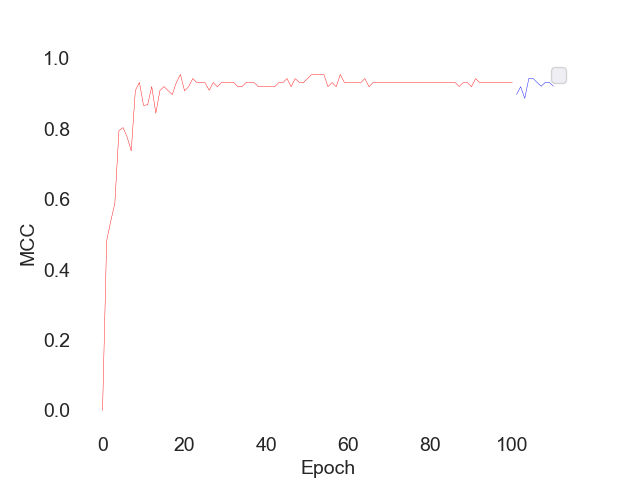}} 
    \subfigure[Credit Approval]{\includegraphics[width=0.3\linewidth]{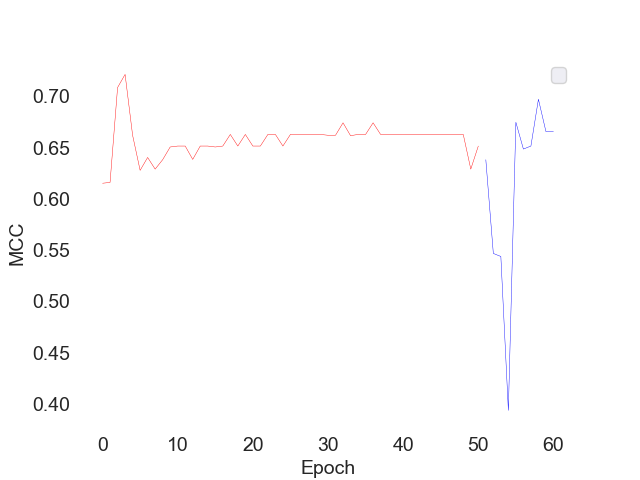}} \\
    \subfigure[Libras]{\includegraphics[width=0.3\linewidth]{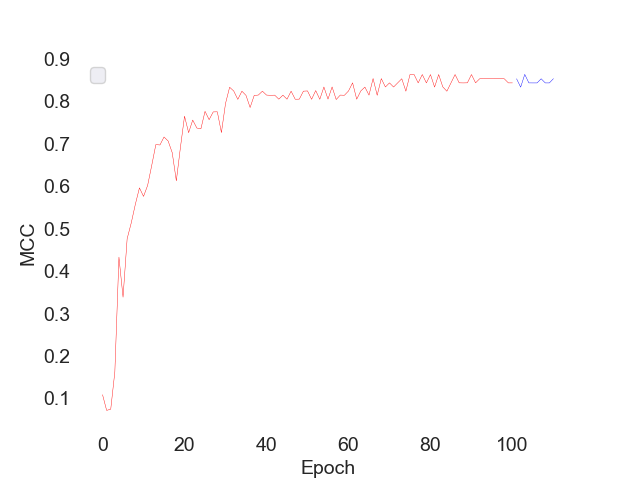}} 
    \subfigure[Cylinder Bands]{\includegraphics[width=0.3\linewidth]{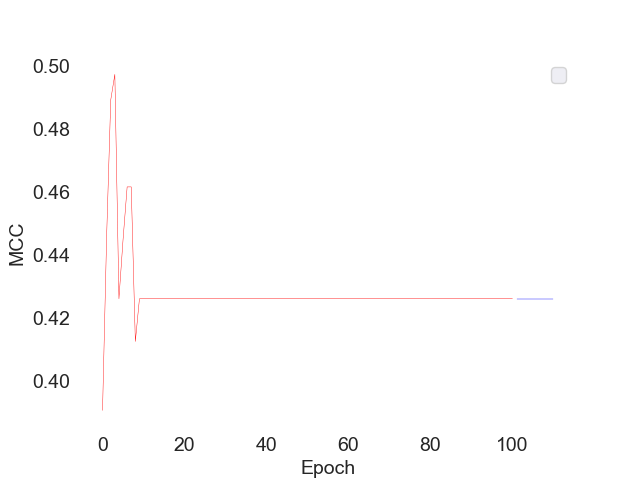}} 
    \caption{Learning curves across training epochs of \our{}. Red color indicates the phase of training adaptation and classification networks while blue color shows the fine-tuning phase of the entire model (including Vit encoder).}
    \label{fig:fine}
\end{figure*}

Figure \ref{fig:fine} presents MCC scores across training epochs for 5 datasets. Red color indicates the phase of training adaptation and classification networks while blue color shows the fine-tuning phase of the entire model (including Vit encoder). As can be seen the learning rate has to be carefully scheduled to avoid drops in performance.

\end{document}